\documentclass[sigconf]{acmart}
\AtBeginDocument{%
  }

\setcopyright{acmcopyright}
\copyrightyear{2022}
\acmYear{2022}
\setcopyright{rightsretained}
\acmConference[KDD '22]{Proceedings of the 28th ACM SIGKDD Conference on Knowledge Discovery and Data Mining}{August 14--18, 2022}{Washington, DC, USA}
\acmBooktitle{Proceedings of the 28th ACM SIGKDD Conference on Knowledge Discovery and Data Mining (KDD '22), August 14--18, 2022, Washington, DC, USA}
\acmDOI{10.1145/3534678.3539364}
\acmISBN{978-1-4503-9385-0/22/08}

\newcommand{\ours}{\texttt{MF-HNP}}
\newcommand{\out}[1]{}
\usepackage{algorithm}
\usepackage{algorithmic}
\usepackage{amsmath,amsfonts,bm}

\def\eqref#1{equation~\ref{#1}}
\def\Eqref#1{Equation~\ref{#1}}
\def\1{\bm{1}}
\DeclareMathAlphabet{\mathsfit}{\encodingdefault}{\sfdefault}{m}{sl}
\SetMathAlphabet{\mathsfit}{bold}{\encodingdefault}{\sfdefault}{bx}{n}

\usepackage{url}

\usepackage{microtype}
\usepackage{graphicx}
\usepackage{booktabs} %
\usepackage{pifont}
\usepackage{hyperref}

\usepackage{caption}
\usepackage{natbib}

\usepackage{amsmath,amssymb,amsfonts,amsthm}

\usepackage{xcolor}
\usepackage{mathtools}
\usepackage{dsfont}
\usepackage{hyperref}
\usepackage{bm}
\usepackage{nicefrac}
\usepackage{wrapfig}
\usepackage{lipsum}

\usepackage{algorithm,algorithmic}
\usepackage[linesnumbered, ruled,vlined,algo2e]{algorithm2e}

\newcounter{assumption}%
\renewcommand{\theassumption}{\arabic{assumption}}

\usepackage{multirow}
\usepackage{caption}
\usepackage{subcaption}

\usepackage{booktabs}
\usepackage{amssymb,amsmath}

\begin{document}

\title{Multi-fidelity Hierarchical Neural Processes}

\author{Dongxia Wu}
\affiliation{%
  \institution{University of California, San Diego}
  \city{La Jolla}
  \state{CA}
  \country{USA}
}
\email{dowu@ucsd.edu}

\author{Matteo Chinazzi}
\affiliation{%
 \institution{Northeastern University}
 \city{Boston}
 \state{MA}
 \country{USA}
 }
 \email{m.chinazzi@northeastern.edu}

\author{Alessandro Vespignani}
\affiliation{%
  \institution{Northeastern University}
  \city{Boston}
  \state{MA}
  \country{USA}
 }
 \email{a.vespignani@northeastern.edu}

\author{Yi-An Ma}
\affiliation{
  \institution{University of California, San Diego}
  \city{La Jolla}
  \state{CA}
  \country{USA}
}
\email{yianma@ucsd.edu}

\author{Rose Yu}
\affiliation{
  \institution{University of California, San Diego}
  \city{La Jolla}
  \state{CA}
  \country{USA}
}
\email{roseyu@ucsd.edu}
\renewcommand{\shortauthors}{Wu et al.}

\begin{abstract}

%such as epidemic models
Science and engineering fields use computer simulation extensively. These simulations are often run at multiple levels of sophistication to balance accuracy and efficiency. Multi-fidelity surrogate modeling reduces the computational cost by fusing different simulation outputs. Cheap data generated from low-fidelity simulators can be combined with limited high-quality data generated by an expensive high-fidelity simulator. Existing methods based on Gaussian processes rely on strong assumptions of the kernel functions and can hardly scale to high-dimensional settings. We propose Multi-fidelity Hierarchical Neural Processes (\ours{}), a unified neural latent variable model for multi-fidelity surrogate modeling. \ours{} inherits the flexibility and scalability of Neural Processes. The latent variables transform the correlations among different fidelity levels from observations to latent space. The predictions across fidelities are conditionally independent given the latent states. It helps alleviate the error propagation issue in existing methods. \ours{} is flexible enough to handle non-nested high dimensional data at different fidelity levels with varying input and output dimensions. We evaluate \ours{} on {epidemiology and climate modeling tasks}, achieving competitive performance in terms of accuracy and uncertainty estimation. In contrast to deep Gaussian Processes \citep{cutajar2019deep} with only low-dimensional ($<$ 10) tasks, our method shows great promise for speeding up high-dimensional complex simulations (over $7,000$ for epidemiology modeling and $45,000$ for climate modeling). 
\end{abstract}



\maketitle

\section{Introduction}
In scientific and engineering applications, a computational model, often realized by simulation, characterizes the input-output relationship of a physical system. The input describes the properties and environmental conditions, and the output describes the quantities of interest. For example, in epidemiology, computational models have long been used to forecast the evolution of epidemic outbreaks and to simulate the effects of public policy interventions on the epidemic trajectory \cite{Halloran4639, Lofgren18095,cramer2021evaluation}. In the case of COVID-19 \cite{chinazzi2020effect,davis2021cryptic}, model inputs range across virus and disease characteristics (e.g. transmissibility and severity), non-pharmaceutical interventions (e.g. travel bans, school closures, business closures), and individual behavioral responses (e.g. changes in mobility and contact rates); while the output describes the evolution of the pandemic (e.g. the time series of the prevalence and incidence of the virus in the population).

Computational models can be simulated at multiple levels of sophistication. High-fidelity models produce accurate output at a higher cost, whereas low-fidelity models generate less accurate output at a cheaper cost. To balance the trade-off between computational efficiency and prediction accuracy, multi-fidelity modeling \cite{peherstorfer2018survey} aims to learn a surrogate model that combines simulation outputs at multiple fidelity levels to accelerate learning. Therefore, we can obtain predictions and uncertainty analysis at high fidelity while leveraging cheap low-fidelity simulations for speedup. 

Since the pioneering work of Kennedy and Hagan \cite{kennedy2000predicting} on modeling oil reservoir simulator, Gaussian processes (GPs) \cite{rasmussen2003gaussian} have become the predominant tools in multi-fidelity modeling. GPs effectively serve as surrogate models to emulate the output distribution of complex physical systems with uncertainty \cite{le2014recursive,perdikaris2016multifidelity,wang2021multi}. However, GPs often struggle with high-dimensional data and require prior knowledge for kernel design. Multi-fidelity GPs also require a nested data structure \citep{perdikaris2017nonlinear} and the same input dimension at each fidelity level \citep{cutajar2019deep}, which significantly hinders their applicability in the real world. Therefore, efforts to combine deep learning and GPs have undergone significant growth in the machine learning community \cite{damianou2013deep, raissi2016deep, wilson2016deep, salimbeni2017doubly}. One of the most scalable frameworks of such combinations is Neural processes (NP) \cite{garnelo2018neural, garnelo2018conditional, kim2019attentive}, which is a neural latent variable model.

Unfortunately, existing NP models are mainly designed for single-fidelity data and cannot handle multi-fidelity outputs. While we can train multiple NPs separately, one for each fidelity, this approach fails to exploit the relations among multi-fidelity models governed by the same physical process. Furthermore, models with more fidelity levels require more training data, which leads to higher computational costs. An alternative is to learn the relationship between low- and high-fidelity model outputs and model the correlation function with NP \cite{wang2020mfpc}.  But this approach always requires paired data at the low- and high-fidelity level. Another limitation is high dimensionality. The correlation function maps from the joint input-output space of the low-fidelity model to the high-fidelity output, which is prone to over-fitting.

In this work, we propose  Multi-Fidelity Hierarchical Neural Process (\ours{}), the first \textit{unified} framework for scalable multi-fidelity modeling in neural processes family. Specifically, \ours{} inherits the properties of Bayesian neural latent variable model while learning the joint distribution of multi-fidelity output. We design a unified evidence lower bound (ELBO) for the joined distribution as a training loss. The code and data are available on
\url{https://github.com/Rose-STL-Lab/Hierarchical-Neural-Processes}.

In summary, our contributions include:
\begin{itemize}
    \item A novel multi-fidelity surrogate model, Multi-fidelity Hierarchical Neural Processes (\ours{}). Its unified framework makes it flexible to fuse data with varying input and output dimensions at different fidelity levels.
    \item  A novel Neural Process architecture with conditional independence at each fidelity level. It fully utilizes the multi-fidelity data,  reduces the input dimension, and alleviates error propagation in forecasting.
    \item Real-world large-scale multi-fidelity application on epidemiology and climate modeling to show competitive accuracy and uncertainty estimation performance.
\end{itemize}

\section{Related Work}
\paragraph{Multi-fidelity Modeling.}
Multi-fidelity surrogate modeling is widely used in science and engineering fields, from climate science \cite{Hosking2020,valero2021multifidelity} to aerospace systems \cite{brevault2020overview}. The pioneering work of \cite{kennedy2000predicting} uses GP to relate models at multiple fidelity levels with an autoregressive model. \cite{le2014recursive} proposed recursive GP with a nested structure in the input domain for fast inference.  \cite{perdikaris2015multi, perdikaris2016multifidelity} deals with high-dimensional GP settings by taking the Fourier transformation of the kernel function. \citep{perdikaris2017nonlinear} proposed multi-fidelity Gaussian processes  (NARGP) but it assumes a  nested structure in the input domain to enable a sequential training process at each fidelity level.  An extreme case that we include in our experiment is when the data sets at low- and high-fidelity levels are disjoint.  None of the high-fidelity data could be used for training, which is a failure case for NARGP. Additionally, the prediction error of the low-fidelity model will propagate to high-fidelity output and explode as the number of fidelity levels increases. \cite{wang2021multi} proposed a Multi-Fidelity High-Order GP model to speed up the physical simulation. They extended the classical Linear Model of Coregionalization (LMC) to nonlinear case and placed a matrix GP prior on the weight functions. Their method is designed for high-dimensional outputs rather than both high-dimensional inputs and outputs.
Deep Gaussian processes (DGP)  \citep{cutajar2019deep} designs a single objective to optimize the kernel parameters at each fidelity level jointly. However, the DGP architecture introduces a constraint that requires the inputs at each fidelity level to be defined by the same domain with the same dimension. Moreover, DGP is still based on GPs, which are not scalable for applications with high-dimensional data. In contrast,  NP is flexible and much more scalable. 
% multi-fidelity modeling application \citep{park2017remarks,de2020transfer,hebbal2021multi}

Deep learning has been applied to multi-fidelity modeling. For example, \cite{guo2022multi} uses deep neural networks to combine parameter-dependent output quantities. \cite{meng2020composite} propose a composite neural network for multi-fidelity data from inverse PDE problems. \cite{meng2021multi} propose Bayesian neural nets for multi-fidelity modeling. \cite{de2020transfer} use transfer learning to fine-tune the high-fidelity surrogate model with the deep neural network trained with low-fidelity data.    \cite{cutajar2019deep,hebbal2021multi} propose deep Gaussian process to capture nonlinear correlations between
fidelities, but their method cannot handle the case where different fidelities have data with different dimensions. 
Tangentially, multi-fidelity methods have also recently been
investigated in Bayesian optimization, active learning, and bandit problems \citep{li2020multi,li2021batch,perry2019allocation,kandasamy2017multi}.

\paragraph{Neural Processes.} Neural Processes (NPs)  \cite{garnelo2018conditional, kim2018attentive, louizos2019functional,singh2019sequential} provide scalable and expressive alternatives to GPs for modeling stochastic processes. However, none of the existing NP models can efficiently incorporate multi-fidelity data.  Earlier work by \cite{raissi2016deep} combines multi-fidelity GP with deep learning by placing a GP prior on the features learned by deep neural networks. However, their model remains closer to GPs. Quite recently, \cite{wang2020mfpc} proposed multi-fidelity neural process with physics constraints (MFPC-Net). They use NP to learn the correlation between multi-fidelity data by mapping both the input and output of the low-fidelity model to the high-fidelity model output. But their model requires paired data and cannot utilize the remaining unpaired data at the low-fidelity level.

% \ry{allen add more neural process papers}

\section{Background}

\begin{figure*}[t!]
    \centering
    \includegraphics[width=0.7\linewidth,trim={100 30 100 30}]{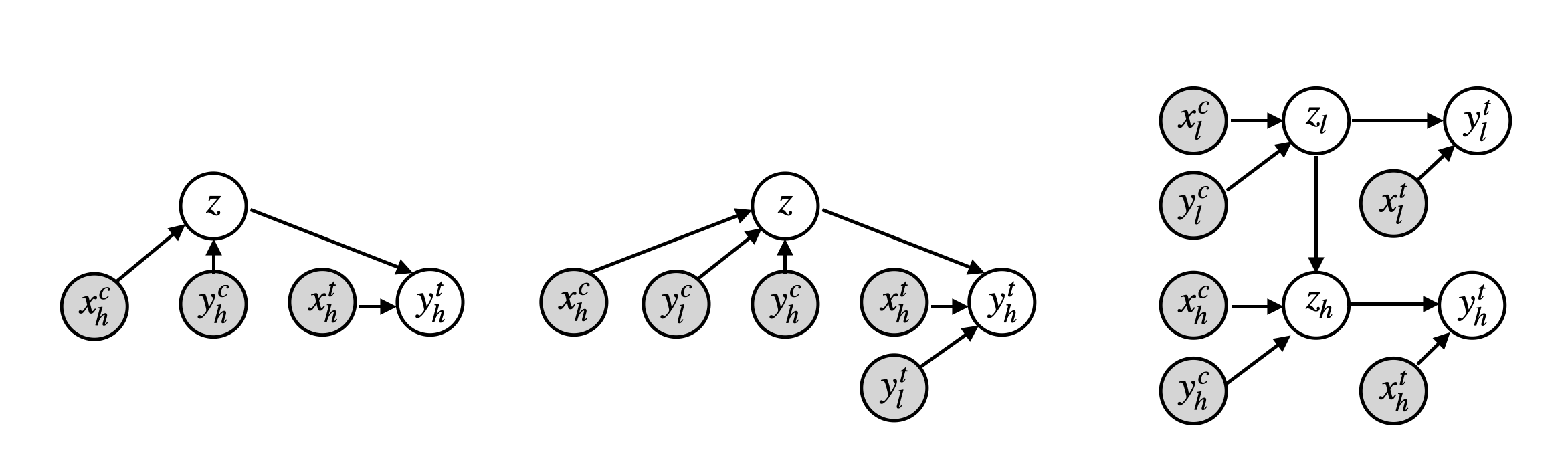}
    \caption{Graphical models for Single-Fidelity Neural Process (left), Multi-Fidelity Neural Process (middle), Multi-Fidelity Hierarchical Neural Process (right). Shaded circles denote observed variables and hollow circle represent latent variables. The directed edges represent conditional dependence. }
    \label{fig:mnp}
\end{figure*}

\subsection{Muti-Fidelity Modeling}
Formally, given input domain $  \mathcal{X} \subseteq  \mathbb{R}^{d_x}$ and output domain $ \mathcal{Y} \subseteq  \mathbb{R}^{d_y}$,   a model is a (stochastic) function $f:\mathcal{X} \rightarrow \mathcal{Y}$.  Evaluations of $f$ incur computational costs $c> 0$. The computational costs $c$ are much higher at higher fidelity level. Therefore, we assume that a limited amount of expensive high-fidelity data is available for training. In multi-fidelity modeling, we have a set of functions $\{f_{1},\cdots, f_{K}\}$ that approximate $f$ with increasing accuracy and computational cost. We aim to learn a surrogate model $\hat{f}_{K}$ that combines information from low-fidelity models with a small amount of data from high-fidelity models.

Given parameters $x_k$ at fidelity level $k$, we query the simulator to obtain data set from different scenarios $\mathcal{D}_k \equiv \{x_{k,i},[{y_{k,i}}]_{s=1}^S\}_i$, where $[{y_{k,i}}]_{s=1}^S$ are $S$ samples generated by $f_{k}(x_{k,i})$ for scenario $i$. In epidemic modeling, for example, each scenario corresponds to a different effective reproduction number of the virus, contact rates between individuals, or the effects of policy interventions. For each scenario, we simulate multiple epidemic trajectories as samples from the stochastic function. We aim to learn a deep surrogate model that approximates the data distribution $p(y^t_K|x^t_K,\mathcal{D}_1^c,\mathcal{D}_2^c, ..., \mathcal{D}_K^c)$ at the highest fidelity level $K$ over the target set $y^t_K$, given context sets at different fidelity levels $\mathcal{D}_k^c  \subset \mathcal{D}_k$ and the corresponding $x^t_K$.

% from new input parameters to outputs at the highest fidelity level, which is 
% $p(y_t^K|x_t^K, (x^1,[{y^1}]_{m=1}^M),  (x^2,[{y^2}]_{m=1}^M),...,  (x^K,[{y^K}]_{m=1}^M))$. 

For simplicity, we use two levels of fidelity, but our framework can be generalized easily. Let us denote the low-fidelity data as $\mathcal{D}_l \equiv \{x_{l,i},[{y_{l,i}}]_{s=1}^S\}_i$  and high-fidelity data as $\mathcal{D}_h \equiv \{x_{h,i},[{y_{h,i}}]_{s=1}^S\}_i$. If $\mathcal{D}_h \subset \mathcal{D}_l$, the data domain has the nested structure. If $\mathcal{D}_h = \mathcal{D}_l$, we say the low- and high-fidelity data sets are paired. Low-fidelity data can be split into context sets $\mathcal{D}_l^c \equiv \{x^c_{l,n},[{y^c_{l,n}}]_{s=1}^S\}_{n=1}^{N_l}$ and target sets $\mathcal{D}_l^t \equiv \{x^t_{l,m},[{y^t_{l,m}}]_{s=1}^S\}_{m=1}^{M_l}$. Similarly, high-fidelity data can be split into context sets $\mathcal{D}_h^c \equiv \{x^c_{h,n},[{y^c_{h,n}}]_{s=1}^S\}_{n=1}^{N_h}$ and target sets $\mathcal{D}_h^t \equiv \{x^t_{h,m},[{y^t_{h,m}}]_{s=1}^S\}_{m=1}^{M_h}$. 

\subsection{Neural Processes}
\label{Background:neural_processes}
%
% \textbf{General Setup.}
%
Neural processes (NPs) \citep{garnelo2018neural} are the family of conditional latent variable models for implicit stochastic processes ($\mathcal{SP}s$) \citep{wang2020doubly}. NPs are in between GPs and neural networks (NNs). Like GPs, NPs can represent distributions over functions and estimate the uncertainty of the predictions. But they are more scalable in high dimensions and can easily adapt to new observations. According to Kolmogorov Extension Theorem \citep{oksendal2003stochastic}, NPs meet exchangeability and consistency conditions to define $\mathcal{SP}s$. Formally, NP includes local latent variables $z \in \mathbb{R}^{d_z}$ and global latent variables $\theta$ and is trained by the context set $\mathcal{D}^c \equiv \{x^c_{n},[{y^c_{n}}]_{s=1}^S\}_{n=1}^{N}$ and target sets $\mathcal{D}^t \equiv \{x^t_{m},[{y^t_{m}}]_{s=1}^S\}_{m=1}^{M}$. Learning the posterior of $z$ and $\theta$ is equivalent to maximizing the following posterior likelihood:

\vskip -0.12in
\begin{align}
    & \prod^{S}_{s=1} p(y^t_{s,1:M}|x^t_{1:M},\mathcal{D}^c,\theta) =\nonumber\\ 
    & \prod^{S}_{s=1}\int p(z_s|\mathcal{D}^c,\theta)\prod^{M}_{m=1}p(y^t_{s,m}|z_s, x^t_{m},\theta)dz_s\nonumber
    % \label{eqn:np}
\end{align}
% \vskip -0.12in

We omit the sample index $s$ in what follows.

\textbf{Approximate Inference.}
Since marginalizing over the local latent variables $z$ is intractable, the NP family \citep{garnelo2018neural, kim2019attentive} introduces approximate inference on latent variables and derives the corresponding evidence lower bound (ELBO) for the training process.

\vskip -0.12in
\begin{align}
% \begin{split}
    & \log p(y^t_{1:M}|x^t_{1:M},\mathcal{D}^c,\theta) \geq \nonumber\\ 
    & \mathbb{E}_{q_\phi(z|\mathcal{D}^c \cup \mathcal{D}^t)} \big[  \sum_{m=1}^M\log p(y^t_m|z, x^t_m,\theta)+\log\frac{q_\phi(z|\mathcal{D}^c)}{q_\phi(z|\mathcal{D}^c \cup \mathcal{D}^t)}\big] \nonumber
    % \label{eqn:elbo_np}
% \end{split}
\end{align}
% \vskip -0.12in

Note that this variational approach approximates the intractable true posterior $p(z|\mathcal{D}^c, \theta)$ with the approximate posterior $q_\phi(z|\mathcal{D}^c)$ . This approach is also an amortized inference method as the global parameters $\phi$ are shared by all context data points. It is efficient during the test time (no per-data-point optimization) \citep{volpp2020bayesian}. 

NPs use NNs to represent $q_\phi(z|\mathcal{D}^c)$, 
and $p(y^t_m|z, x^t_m,\theta)$. $q_\phi()$ is referred as the encoder network ($\mathrm{Enc}$, determined by the parameters $\phi$). $p(.|\theta)$ is referred as the decoder network ($\mathrm{Dec}$, determined by parameters $\theta$). These two networks assume that the latent variable $z$ and the outputs $y$ follow the factorized Gaussian distribution determined by mean and variance.

\vskip -0.17in
\begin{align}
% \begin{split}
    & q_\phi(z|\mathcal{D}^c) = \mathcal{N}(z|\mu_z,\mathrm{diag}(\sigma^2_z)) \nonumber\\ 
    & \mu_z=\mathrm{Enc}_{\mu_z,\phi}(\mathcal{D}^c), \:\:\: \sigma_z^2=\mathrm{Enc}_{\sigma_z^2,\phi}(\mathcal{D}^c) \nonumber\\ 
    & p(y^t_m|z,x^t_m,\theta) = \mathcal{N}(y^t_m|\mu_y,\mathrm{diag}(\sigma^2_y)) \nonumber\\ 
    & \mu_y=\mathrm{Dec}_{\mu_y,\theta}(z,x^t_m), \:\:\: \sigma_y^2=\mathrm{Dec}_{\sigma_y^2,\theta}(z,x^t_m) \nonumber 
% \label{eqn:encdec_np1}
% \end{split}
\end{align}
% \vskip -0.1in

\textbf{Context Aggregation.}
  Context aggregation aggregates all context points $\mathcal{D}^c$ to infer latent variables $z$. To meet the exchangeability condition, the context information acquired by NPs should be invariant to the order of the data points. \citet{garnelo2018conditional, garnelo2018neural, kim2018attentive} use mean aggregation (MA). They map the data pair$(x^c_n, y^c_n)$ to a latent representation $r_n = \mathrm{Enc}_{r,\phi}(x^c_n, y^c_n) \in \mathbb{R}^{d_r}$, then apply the mean operation to the entire set $\{r_n\}^N_{n=1}$ to obtain the aggregated latent representation $\bar{r}$. $\bar{r}$ can be mapped to $\mu_z$ and $\sigma^2_z$ to represent the posterior $q_\phi(z|\mathcal{D}^c)$ with an additional neural network encoder. MA uses two encoder networks. $\mathrm{Enc}_{r,\phi}(x^c_n, y^c_n) \in \mathbb{R}^{d_r}$ maps the data pair$(x^c_n, y^c_n)$ to $r_n$ for context aggregation.  $\mathrm{Enc}_{z,\phi}(\bar{r}) \in \mathbb{R}^{d_z}$ maps $\bar{r}$ to $\mu_z$ and $\sigma^2_z$ for latent parameter inference.

\citet{volpp2020bayesian} proposed Bayesian aggregation (BA), which merges these two steps. They define a probabilistic observation model $p(r|z)$ for $r$ depended on $z$, and update $p(z)$ posterior using the Bayes rule $p(z|r_n) = p(r_n|z)p(z)|p(r_n)$ given latent observation $r_n = \mathrm{Enc}_{r,\phi}(x^c_n,y^c_n)$. The corresponding factorized Gaussian for the inference step:

\vskip -0.17in
\begin{align}
% \begin{split}
    & p(r_n|z) = \mathcal{N}(r_n|z,\mathrm{diag}(\sigma^2_{r_n})) \nonumber\\ 
    & r_n=\mathrm{Enc}_{r,\phi}(x^c_n,y^c_n)\nonumber\\  
    & \sigma_{r_n}^2=\mathrm{Enc}_{\sigma_{r_n}^2,\phi}(x^c_n,y^c_n)  \nonumber 
    % \label{eqn:enc_ba_np}
% \end{split}
\end{align}
% \vskip -0.12in

They use a factorized Gaussian prior $p_0(z) \equiv \mathcal{N}(z|\mu_{z,0},\mathrm{diag}(\sigma^2_{z,0}))$ to derive the parameters of posterior $q_\phi(z|\mathcal{D}^c)$:

\vskip -0.12in
\begin{align}
% \begin{split}
    & \sigma^2_z = \big[(\sigma^2_{z,0})^\ominus + \sum^N_{n=1}(\sigma^2_{r_n})^\ominus)\big]^\ominus, \nonumber\\ 
    & \mu_z = \mu_{z,0} + \sigma^2_{z}\odot \sum^N_{n=1}(r_n - \mu_{z,0}) \oslash (\sigma^2_{r_n}). \nonumber 
    % \label{eqn:ba_post_z}
% \end{split}
\end{align}
% \vskip -0.12in

Compared with MA, which treats every context sample equally, BA uses observation variance $\sigma^2_{r_n}$ to weigh the importance of each latent representation $r_n$. BA also represents a permutation-invariant operation on $\mathcal{D}^c$.

\section{Methodology}

In this section, we introduce our proposed Multi-fidelity Hierarchical Neural Processes (\ours) model in three subsections.
The first section discusses the unique architecture of hierarchical neural processes for the multi-fidelity problem. 
Then, we develop the corresponding approximate inference method with a unified ELBO. Finally, we introduce $3$ ELBO variants for scalable training.

\begin{table*}[t]
\begin{center}
\begin{sc}
\caption{Comparison of different NP models at high-fidelity level.}
\label{tab:summary-table}
\begin{tabular}{cccc}
\toprule
Neural Processes Family & Prior Distribution & Posterior Distribution & Generative model \\
\midrule
SF-NP \cite{garnelo2018neural} & $q(z_h|\mathcal{D}^c_h)$ & $p(z|\mathcal{D}^c_h, \mathcal{D}^t_h)$ & $p(y^t_h|x^t_h,z)$\\
MF-NP \cite{wang2020mfpc} & $q(z_h|\mathcal{D}^c_h)$ & $p(z|\mathcal{D}^c_h, \mathcal{D}^t_h)$ & $p(y^t_h|x^t_h, y^t_l, z)$\\
MF-HNP(as)  & $q(z_h|z^{(s)}_l,\mathcal{D}^c_h)$ & $p(z_h|z^{(s)}_l, \mathcal{D}^c_h, \mathcal{D}^t_h)$ & $p(y^t_h|x^t_h,z_h)$\\
MF-HNP(mean) & $q(z_h|\mu_{z_l}, \mathcal{D}^c_h)$ & $p(z_h|\mu_{z_l}, \mathcal{D}^c_h, \mathcal{D}^t_h)$ & $p(y^t_h|x^t_h,z_h)$\\
MF-HNP(mean,std)    & $q(z_h|\mu_{z_l}, \sigma_{z_l}, \mathcal{D}^c_h)$ & $p(z_h|\mu_{z_l}, \sigma_{z_l}, \mathcal{D}^c_h, \mathcal{D}^t_h)$ & $p(y^t_h|x^t_h,z_h)$\\
\bottomrule
\end{tabular}
\end{sc}
\end{center}
\end{table*}

\subsection{Multi-fidelity Hierarchical Neural Processes}
Our high-level goal is to train a deep surrogate model to mimic the behavior of a complex stochastic simulator at the highest level of fidelity. 
% We propose the \ours{} model inspired by deep Gaussian processes (DGP)  \citep{cutajar2019deep}. 
\ours{} inherits the properties of  Bayesian neural latent variable model while learning the joint distribution of multi-fidelity output. It adopts a single objective function for multi-fidelity training. It reduces the input dimension and alleviates error propagation by introducing the hierarchical structure in the dependency graph. 

Figure \ref{fig:mnp} compares the graphical model of \ours{} with Multi-fidelity Neural Process (MF-NP) \cite{wang2020mfpc} and Single-Fidelity Neural Process (SF-NP).  SF-NP assumes that the high-fidelity data is independent of the low-fidelity data and reduces the model to vanilla NP setting. Details of SF-NP and MF-NP are shown in Appendix \ref{Appendix:np_baseline}. \ours{} assignes latent variables $z_l$ and $z_h$ at each fidelity level. The prior of $z_h$ is conditioned on $z_l$, parameterized by a neural network. We use Monte Carlo (MC) sampling method to approximate the posterior of $z_l$ and $z_h$ to calculate the ELBO. 
% Details are discussed in \ref{Method:scalable_training}. 

One key feature of \ours{} is that the model outputs at each fidelity level are conditionally independent given the corresponding latent state. This design transforms the correlations between fidelity levels from the input and output space to the latent space. Specifically, compared with MF-NP where $\hat{y}_h$ depends on $(x_h,y_l)$ input pairs given $z$, $\hat{y}_h$ only depends on input $x_h$ given $z_h$ in \ours{}. It helps \ours{} to significantly reduce the high-fidelity input dimension. In addition, local latent variables at each level of fidelity enable \ours{} to perform both inference and generative modeling separately at each fidelity level. It means \ours can fully utilize the low-fidelity data and is applicable to arbitrary multi-fidelity data sets. As \ours can reduce the input dimension and fully utilize the training data, its prediction performance is significantly improved with limited training data.

Note that in two fidelity setup, \ours{} is related to Doubly Stochastic Variational Neural Process (DSVNP) model proposed by \citet{wang2020doubly} which  introduces local latent variables together with the global latent variables. Different from  DSVNP, \ours{} gives latent variables with separable representations. $z_l$, $z_h$ represent the low- and high-fidelity functional, respectively.

\subsection{Unified ELBO}

We design a unified ELBO as the objective for \ours{}. Unlike vanilla NPs, we need to infer the latent variables $z_l$ and $z_h$ at each fidelity level instead of the global $z$. For the two-fidelity level setup, we use two encoders $q_{\phi_l}(z_l|\mathcal{D}^c_l)$, $q_{\phi_h}(z_h|z_l,\mathcal{D}^c_h)$, and two decoders $p(y^t_l|z_l,x^t_l,\theta_l)$, $p(y^t_h|z_h,x^t_h,\theta_h)$. These four networks approximate the distributions of the latent variables $z_l$, $z_h$ and outputs $y_l$ and $y_h$. Assuming a factorized Gaussian distribution, we can parameterize the distributions by their mean and variance.

\vskip -0.12in
\begin{align}
% \begin{split}
    & q_{\phi_l}(z_l|\mathcal{D}^c_l) = \mathcal{N}(z_l|\mu_{z_l},\mathrm{diag}(\sigma^2_{z_l})) \nonumber\\ 
    & q_{\phi_h}(z_h|z_l,\mathcal{D}^c_h) = \mathcal{N}(z_h|\mu_{z_h},\mathrm{diag}(\sigma^2_{z_h})) \nonumber\\ 
    & p(y^t_{l,m}|z_l,x^t_{l,m},\theta_l) = \mathcal{N}(y^t_{l,m}|\mu_{l,m},\mathrm{diag}(\sigma^2_{y_l})) \nonumber\\ 
    & p(y^t_{h,m}|z_h,x^t_{h,m},\theta_h) = \mathcal{N}(y^t_{h,m}|\mu_{h,m},\mathrm{diag}(\sigma^2_{y_h})) \nonumber 
    % \label{eqn:encdec_mfhnp1}
% \end{split}
\end{align}
% \vskip -0.12in

where 
\vskip -0.12in
\begin{align}
% \begin{split}
    & \mu_{z_l}=\mathrm{Enc}_{\mu_{z_l},{\phi_l}}(\mathcal{D}^c_l), \:\:\: \sigma_{z_l}^2=\mathrm{Enc}_{\sigma_{z_l}^2,{\phi_l}}(\mathcal{D}^c_l) \nonumber \\
    & \mu_{z_h}=\mathrm{Enc}_{\mu_{z_h},{\phi_h}}(z_l,\mathcal{D}^c_h), \:\:\: \sigma_{z_h}^2=\mathrm{Enc}_{\sigma_{z_h}^2,{\phi_h}}(z_l,\mathcal{D}^c_h) \nonumber \\
    & \mu_{y_l}=\mathrm{Dec}_{\mu_{y_l},\theta_l}(z_l,x^t_{l,m}), \:\:\: \sigma_{y_l}^2=\mathrm{Dec}_{\sigma_{y_l}^2,\theta_l}(z_l,x^t_{l,m}) \nonumber \\ 
    & \mu_{y_h}=\mathrm{Dec}_{\mu_{y_h},\theta_h}(z_h,x^t_{h,m}), \:\:\: \sigma_{y_h}^2=\mathrm{Dec}_{\sigma_{y_h}^2,\theta_h}(z_h,x^t_{h,m}) \nonumber
    % \label{eqn:encdec_mfhnp2}
% \end{split}
\end{align}
% \vskip -0.12in

We derive the unified ELBO containing these four terms:

\vskip -0.12in
\begin{align}
% \begin{split}
    &\log p(y^t_{l},y^t_{l}|x^t_{l},x^t_{h},\mathcal{D}^c_l,\mathcal{D}^c_h,\theta)  \nonumber\\ 
    & \geq \mathbb{E}_{q_\phi(z_l,z_h|\mathcal{D}^c_l \cup \mathcal{D}^t_l,\mathcal{D}^c_h \cup \mathcal{D}^t_h)} \big[\log p(y^t_l,y^t_h|z_l,z_h, x^t_l, x^t_h,\theta) \nonumber  \\
    & +\log \frac{q_\phi(z_l,z_h| \mathcal{D}^c_l, \mathcal{D}^c_h)}
    {q_\phi(z_l,z_h|\mathcal{D}^c_l \cup \mathcal{D}^t_l,\mathcal{D}^c_h \cup \mathcal{D}^t_h)} \big] \nonumber\\
    & = \mathbb{E}_{q_{\phi_h}(z_h|z_l,\mathcal{D}^c_h \cup \mathcal{D}^t_h)q_{\phi_l}(z^l|\mathcal{D}^c_l \cup \mathcal{D}^t_l)}\big[ \log p(y^t_h|z_h, x_h^t,\theta_h)  \nonumber\\
    & + \log p(y_l^t|z_l, x_l^t,\theta_l) + 
    \log \frac{q_{\phi_h}(z_h|z_l,\mathcal{D}^c_h)}{q_{\phi_h}(z_h|z^l,\mathcal{D}^c_h \cup \mathcal{D}^t_h)} \nonumber\\
    &  + \frac{q_{\phi_l}(z_l|\mathcal{D}^c_l)}{q_{\phi_l}(z_l|\mathcal{D}^c_l \cup \mathcal{D}^t_l)}\big]
    \label{eqn:mfhnp_elbo}
\end{align}
% \vskip -0.1in

The derivation is based on the conditional independence of \ours architecture shown in Figure \ref{fig:mnp}.

\subsection{Scalable Training}
\label{Method:scalable_training}
To calculate the ELBO in \Eqref{eqn:mfhnp_elbo} for the proposed \ours{} model, we use Monte Carlo (MC) sampling to optimize the following objective function:
\vskip -0.12in
\begin{align}
% \begin{split}
    \mathcal{L}_{MC} & = \frac{1}{K}\sum^{K}_{k=1}\big[\frac{1}{S}\sum^{S}_{s=1}\log p(y_h^t|x_h^t,z_h^{(s)},z_l^{(k)}) \nonumber\\
    & - \text{KL}[q(z_h|z_l^{(k)},\mathcal{D}^c_h, \mathcal{D}^t_h))\|p(z_h|z_l^{(k)},\mathcal{D}^c_h]\big] \nonumber\\
    & + \frac{1}{K}\sum^{K}_{k=1}\log p(y_l^t|x_l^t,z_l^{(k)})  - \text{KL}\big[q(z_l|\mathcal{D}^c_l, \mathcal{D}^t_l)\|p(z_l|\mathcal{D}^c_l)\big] \nonumber
    % \label{eqn:mfhnp_nested}
% \end{split}
\end{align}
% \vskip -0.12in

where the latent variables $z_l^{(k)}$ and $z_h^{(s)}$ are sampled by $q_{\phi_l}(z_l|\mathcal{D}^c_l)$ and $q_{\phi_h}(z_h|z_l^{(k)},\mathcal{D}^c_h)$ respectively.
This standard MC sampling method requires nested sampling. For data sets with multiple fidelity levels, it is computationally challenging. 

An alternative way is to use ancestral sampling \citep{wang2020doubly} (denoted by MF-HNP(AS)) for scalable training and write the estimation as:

\vskip -0.12in
\begin{align}
% \begin{split}
    \mathcal{L}_{AS} & = \frac{1}{S}\sum^{S}_{s=1}\big[\log p(y_h^t|x_h^t,z_h^{(s)},z_l^{(s)}) \nonumber\\
    & - \text{KL}[q(z_h|z_l^{(s)},\mathcal{D}^c_h, \mathcal{D}^t_h))\|p(z_h|z_l^{(s)},\mathcal{D}^c_h]\big] \nonumber\\
    & + \frac{1}{K}\sum^{K}_{k=1}\log p(y_l^t|x_l^t,z_l^{(k)}) - \text{KL}\big[q(z_l|\mathcal{D}^c_l, \mathcal{D}^t_l)\|p(z_l|\mathcal{D}^c_l)\big]
    \label{eqn:mfhnp_as}
% \end{split}
\end{align}
% \vskip -0.12in

We also design two different techniques to infer $z_h$ using either low-level mean of latent variables $\mu_{z_l}$ (denoted by MF-HNP(MEAN)) or both low-level mean and standard deviation $(\mu_{z_l}, \sigma^2_{z_l})$(denoted by MF-HNP(MEAN,STD)). The corresponding ELBOs are:

\vskip -0.12in
\begin{align}
% \begin{split}
    \mathcal{L}_{\mu} & = \frac{1}{S}\sum^{S}_{s=1}\log p(y_h^t|x_h^t,z_h^{(s)},\mu_{z_l}) \nonumber\\
    & - \text{KL}[q(z_h|\mu_{z_l},\mathcal{D}^c_h, \mathcal{D}^t_h))\|p(z_h|\mu_{z_l},\mathcal{D}^c_h] \nonumber\\
    & + \frac{1}{K}\sum^{K}_{k=1}\log p(y_l^t|x_l^t,z_l^{(k)}) - \text{KL}\big[q(z_l|\mathcal{D}^c_l, \mathcal{D}^t_l)\|p(z_l|\mathcal{D}^c_l)\big]
    \label{eqn:mfhnp_mu}
% \end{split}
\end{align}
% \vskip -0.12in

\begin{align}
% \begin{split}
    \mathcal{L}_{\mu,\sigma} & = \frac{1}{S}\sum^{S}_{s=1}\log p(y_h^t|x_h^t,z_h^{(s)},\mu_{z_l},\sigma_{z_l}) \nonumber\\
    & - \text{KL}[q(z_h|\mu_{z_l},\sigma_{z_l},\mathcal{D}^c_h, \mathcal{D}^t_h))\|p(z_h|\mu_{z_l},\sigma_{z_l},\mathcal{D}^c_h] \nonumber\\
    & + \frac{1}{K}\sum^{K}_{k=1}\log p(y_l^t|x_l^t,z_l^{(k)}) - \text{KL}\big[q(z_l|\mathcal{D}^c_l, \mathcal{D}^t_l)\|p(z_l|\mathcal{D}^c_l)\big]
    \label{eqn:mfhnp_mustd}
% \end{split}
\end{align}
% \vskip -0.12in

\begin{figure}[t!]
    \centering
    \includegraphics[width=0.75\linewidth,trim={100 16 100 16}]{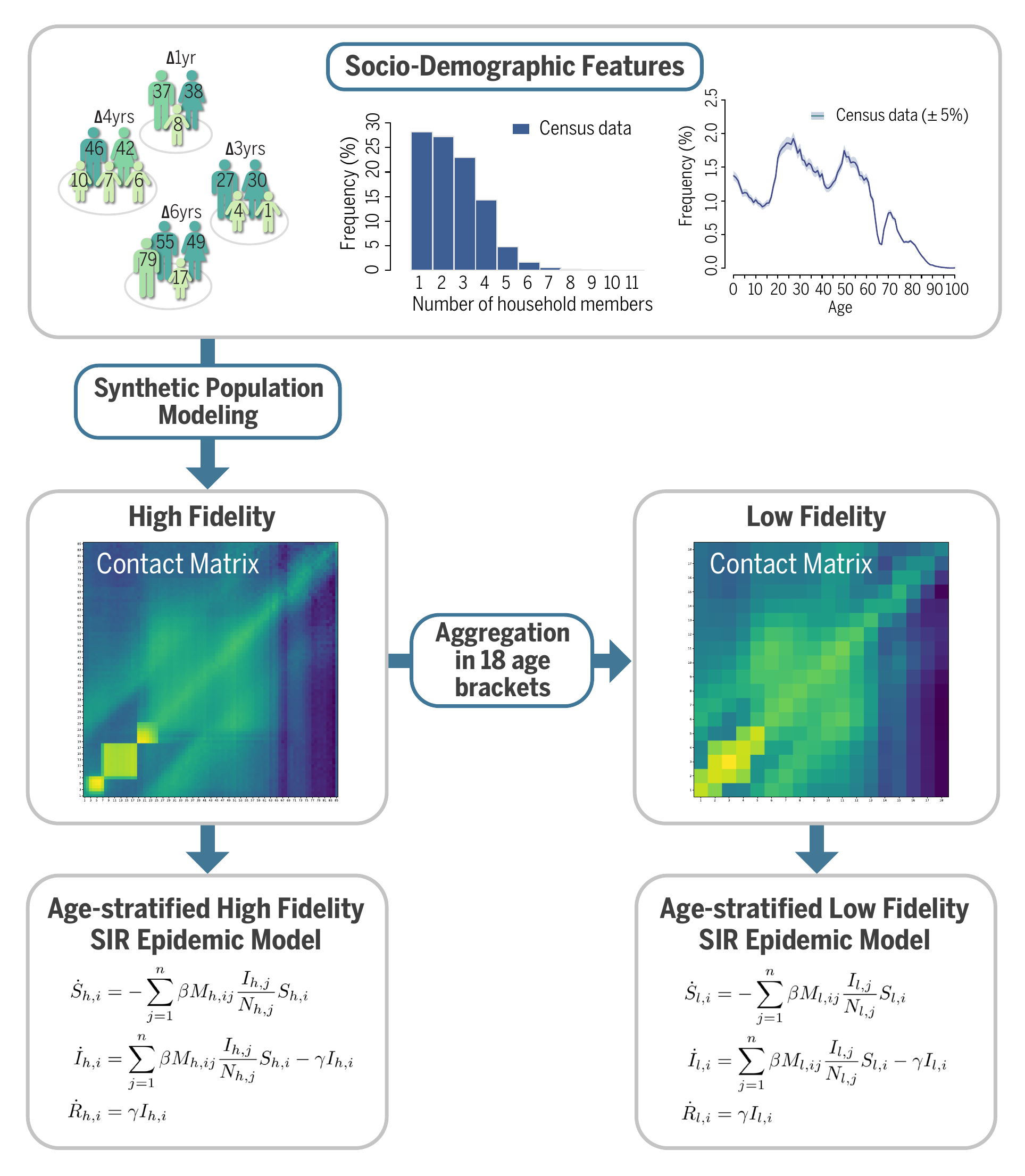}
    \caption{AS-SIR Modeling Framework: First, high-fidelity population-level contact matrices are generated using  macro (census) and micro (survey) data \cite{mistry2021inferring}. Second, low-fidelity contact matrices are obtained by grouping individuals in fewer age brackets. Distinct age-stratified SIR models are used to simulate the epidemic at the two fidelity levels.  
    \label{fig:fig2}
    }
    \vspace{-3mm}
\end{figure}
We include \Eqref{eqn:mfhnp_as},  \Eqref{eqn:mfhnp_mu}, and  \Eqref{eqn:mfhnp_mustd} as the training loss functions for ablation study. The comparison  of different NP models including SF-NP, MF-NP, \ours{} variants for high-fidelity level inference and output generation is shown in Table \ref{tab:summary-table}.

\section{Experiments}
We benchmark the performance of different methods on two multi-fidelity modeling tasks: stochastic epidemiology modeling and climate forecasting. Epidemiology modeling is age-stratified and climate (temperature) modeling is on a regular grid.  

% The fidelity level is based on the number of age brackets and spatial resolution, respectively.

\subsection{Experiment Setup.}
For all experiments, we compare our proposed \ours{} model with both the GP and NP baselines. 

\begin{itemize}
    \item GP baselines include the nonlinear autoregressive multi-fidelity GP regression model (NARGP) \citep{perdikaris2017nonlinear} and single-fidelity Gaussian Processes (SF-GP) which assumes that the data are independent at each fidelity level.
    \item NP baselines include single-fidelity Neural Processes (SF-NP) and multi-fidelity Neural Processes (MF-NP) \citep{wang2020mfpc}.
    \item  For our proposed \ours{} model, we provide $3$ variants to approximate inference for ablation study, including inference by low-level mean of latent variables (MF-HNP(MEAN)), low-level mean and standard deviation of latent variables (MF-HNP(MEAN,STD)), and ancestral sampling method (MF-HNP(AS)). Details have been discussed in Section \ref{Method:scalable_training}.
\end{itemize}

 For NP models, we also consider two different context aggregation methods discussed in Section \ref{Background:neural_processes}, including mean context aggregation and Bayesian context aggregation. Both are applied to generate latent variables $z$ at each fidelity level. For the NARGP and MF-NP baseline, they only work for the data with nested data structure based on their model architecture and assumption \citep{perdikaris2017nonlinear}. For MF-NP, it requires both low-fidelity simulation output $y^l$ and high-fidelity input $x^h$ as model input. Therefore, we assume that $y^l$ is known for the validation and test set for MF-NP, which means MF-NP requires more data compared with \ours and other baselines.

We report the mean absolute error (MAE) for accuracy estimation. For uncertainty estimation, we use mean negative log-likelihood (NLL). For age-stratified Susceptible-Infectious-Recovered (AS-SIR) experiment, we perform a log transformation on the number of infections in the output space to deal with the long-tailed distribution. NLL for AS-SIR experiment is calculated in the log space, while MAE is calculated in the original space. For climate modeling experiment, both NLL and MAE are measured in the original space. We calculate the NLL based on the Gaussian distribution determined by model outputs of mean and standard deviation, and MAE between the mean predictions and the truth.

\begin{figure*}[t]
    \centering
    \includegraphics[width=\linewidth,trim={100 70 100 35}]{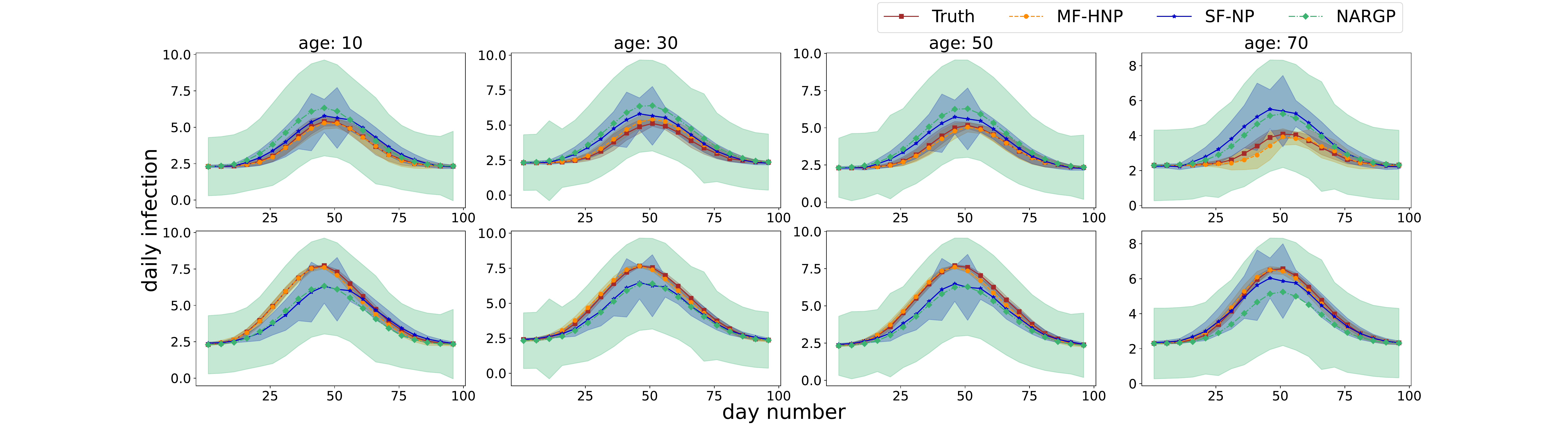}
    \caption{100 days ahead infectious incidence compartment forecasting of randomly selected scenario at each row, analyzed in $4$ age groups. Natural log scale for $y$ axis.}
    \label{fig:sir_pred}
\end{figure*}

\begin{table*}[ht]
\caption{Prediction performance comparison on Age-Stratified SIR data sets.}
\label{tb:sir-table}
% \vskip 0.15in
\begin{center}
\begin{sc}
\begin{tabular}{c|c|c|c|c|c}
\toprule
Data & Method & MAE (nested) $\downarrow$ & NLL (nested) $\downarrow$ & MAE (non-nested) $\downarrow$ & NLL (non-nested) $\downarrow$\\
\midrule
\multirow{12}{*}{nested} & SF-GP & $342.99\pm0.04$ & $\mathbf{1.71}\pm0.06$ & $342.99\pm0.04$ & $\mathbf{1.71}\pm0.06$\\ 
& NARGP & $342.72\pm0.13$ & $1.78\pm0.1$ & $\times$  & $\times$ \\ 
& SF-NP-MA  & $333.41\pm100.73$ & $6.14\pm4.11$ & $333.41\pm100.73$ & $6.14\pm4.11$ \\ 
& MF-NP-MA  & $341.08\pm0.18$ & $6.5\pm0.58$ & $\times$  & $\times$ \\
& MF-HNP(mean)-MA  & $257.39\pm24.17$ & $11.09\pm11.93$  & $249.5\pm25.82$ & $10.58\pm11.33$\\
& MF-HNP(mean,std)-MA  & $257.0\pm23.13$ & $9.26\pm9.38$ & $254.04\pm18.0$ & $13.04\pm14.84$\\
& MF-HNP(as)-MA  & $266.17\pm16.13$ & $10.59\pm11.06$  & $262.61\pm10.68$ & $11.66\pm12.71$\\
& SF-NP-BA  & $294.3\pm75.81$ & $36.35\pm46.5$  & $294.3\pm75.81$ & $36.35\pm46.5$ \\ 
& MF-NP-BA  & $340.22\pm1.51$ & $4.34\pm2.23$  & $\times$  & $\times$ \\
& MF-HNP(mean)-BA  & $\mathbf{201.56}\pm61.15$ & $1.97\pm0.44$  & $\mathbf{199.75}\pm64.51$ & $1.95\pm0.5$\\
& MF-HNP(mean,std)-BA  & $229.09\pm77.44$ & $8.24\pm9.54$  & $203.05\pm65.84$ & $6.66\pm7.19$\\
& MF-HNP(as)-BA  & $205.26\pm49.1$ & $2.69\pm1.0$  & $205.43\pm43.79$ & $3.24\pm1.59$\\
\bottomrule
\end{tabular}
\end{sc}
\end{center}
\vskip -0.1in
\end{table*}
\subsection{Age-Stratified SIR Compartmental Model}

We use an age-stratified Susceptible-Infectious-Recovered (AS-SIR) epidemic model: 
$$ \Dot{S}_i = -\lambda_i S_i, \:\:\: \Dot{I}_i = \lambda_i S_i - \gamma I_i, \:\:\: \Dot{R}_i = \gamma I_i $$
where $S_i$, $I_i$, and $R_i$ denote the number of susceptible, infected, and recovered individuals of age $i$, respectively. The age-specific force of infection is defined by $\lambda_i$ and it is equal to:

$$ \lambda_i = \beta \sum_j M_{i,j} \frac{I_j}{N_j}, $$

where $\beta$ denotes the transmissibility rate of the infection, $N_j$ is the total number of individuals of age $j$, and $M_{i,j}$ is the overall age-stratified contact matrices describing the average number of contacts with individuals of age $j$ for an individual of age $i$.

This model assumes heterogeneous mixing between age groups, where the population-level contact matrices $M$ are generated using highly detailed macro (census) and micro (survey) data on key socio-demographic features \cite{mistry2021inferring} to realistically capture the social mixing differences that exist between different countries/regions of the world and that will affect the spread of the virus.

\textbf{Dataset.}
We include overall $109$ scenarios at different locations in China, U.S., Europe. The data in China is at the province level. The data in the U.S. is at state level. The data in Europe is at the country level. For each scenario, we generate $30$ samples for 100 day's new infection prediction at low- and high-fidelity levels based on the corresponding initial conditions, $R_0$, age-stratified population, and the overall age-stratified contact matrices. The high-fidelity data, as shown in Figure \ref{fig:fig2}, has $85$ age groups. The size of the age-stratified contact matrices $M_{h,ij}$ is $85\times85$. For low-fidelity data, we aggregate the data and obtain 18 age groups, resulting in a contact matrix $M_{l,ij}$ of size $18\times18$.

We randomly split 31 scenarios for training candidate set, $26$ scenarios for the validation set and $52$ scenarios for test set at both fidelity levels. In the nested data set case, we first randomly select $26$ scenarios from the training candidate set as the training set at low-fidelity level, then randomly select $5$ scenarios from them as the training set at high-fidelity level. In the non-nested data set case, we randomly split $26$ scenarios as the training set at low-fidelity level and $5$ scenarios as the training set at high-fidelity level. The validation and test set are both at high-fidelity level.

\textbf{Performance Analysis.}
Table \ref{tb:sir-table} compares the prediction performance for 2 GP methods and 10 NP methods for 100 day ahead infection forecasting. The performance is reported in MAE and NLL over 100 days. \ours{}(MEAN)-BA has the best prediction performance in terms of MAE for both the scenario with nested data structure and non-nested data structure. GP baselines SF-GP and NARGP have similar worst MAE, which means the low-fidelity data does not help NARGP learn useful information. Because in high-dimensions, the strict assumption of no observation noise at low-fidelity level does not hold for NARGP. 

For NP baselines, MF-NP-(MA/BA) baselines have worse accuracy performance compared with the SF-NP-(MA/BA) baselines. This is due to the limited number of paired training data that MF-NP can utilize. The small number of training data plus the high-dimensional input and output space makes it difficult for MF-NP to learn the correct pattern for model predictions. For all NP models, we find Bayesian aggregation improves the performance. With respect to different hierarchical inference methods of \ours{}. Table \ref{tb:sir-table} shows \ours(AS) and \ours(MEAN) have superior performance compared to \ours(MEAN,STD) in terms of both NLL and MAE.

Figure \ref{fig:sir_pred} visualizes the prediction results of two randomly selected scenarios in the nested dataset. It shows the truth, our \ours{} prediction together with two other baselines representing the best GP baseline and the best NP baseline in four age groups (10,30,50,70). In this experiment, the best GP is NARGP and the best NP is SF-NP. 
% \ours(MEAN) is the most accurrate with respect to MAE, while \ours(MC) has similar performance but smaller standard deviation.Also, both of them have small NLL score. 
One interesting finding is that although SF-GP has the best NLL score, the visualization  shows its prediction is very conservative  by generating a large confidence interval, which is not informative. On the contrary, \ours prediction is able to generate a narrower confidence interval while covering the truth at the same time (shown in Figure \ref{fig:sir_pred}). 

When switching to non-nest data set, the \ours{} model is still reliable for this much harder task. In fact, the MAE performance of \ours(MEAN)-BA is even better.

\subsection{Climate Model for Temperature.}
\begin{figure}[h]
    \centering
    \includegraphics[width=0.95\linewidth]{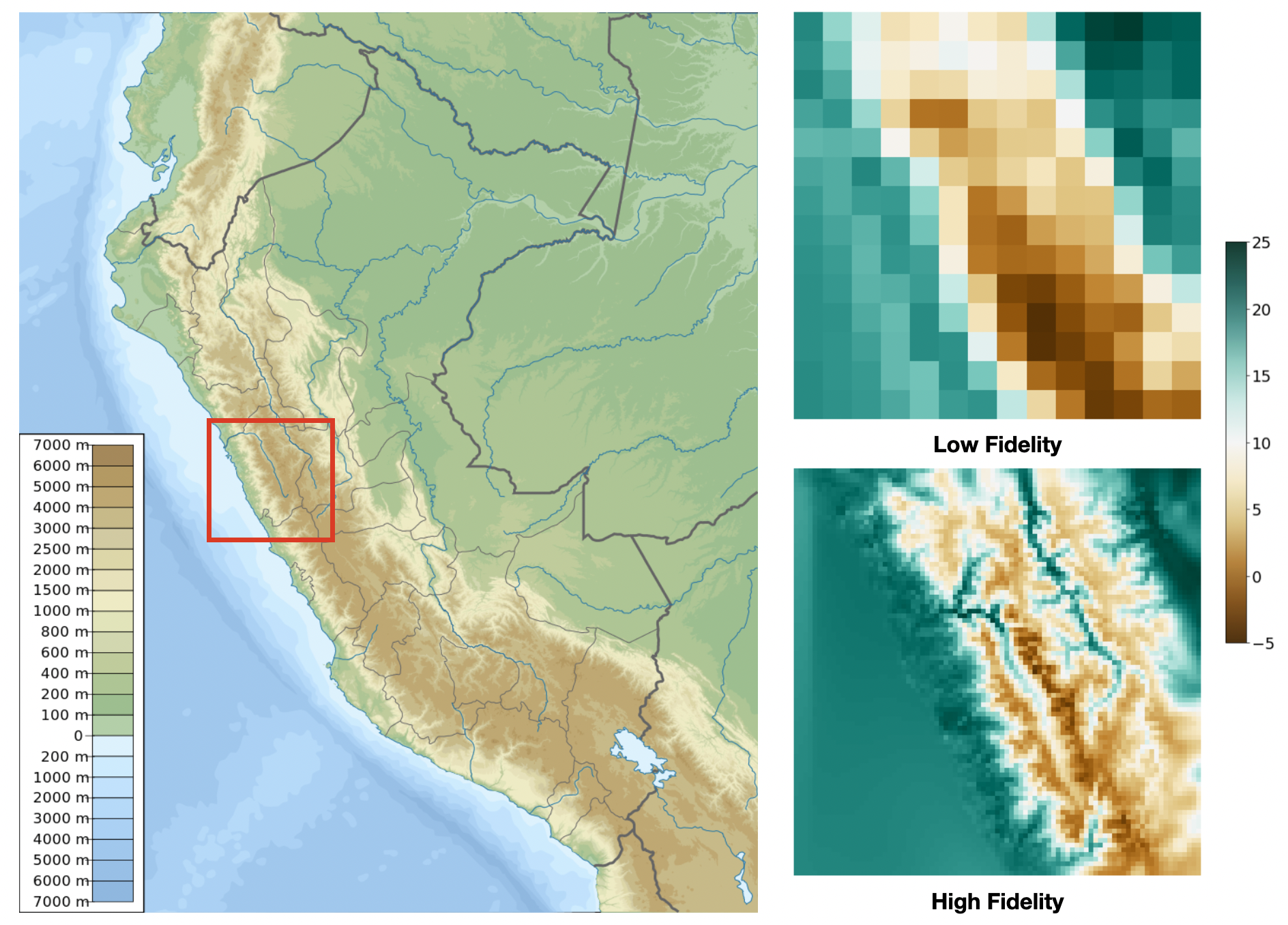}
    \caption{Left: Region of interest \citep{wiki_peru}. Upper Right: sample from low-fidelity temperature model. Lower Right: sample from high-fidelity temperature model.}
    \label{fig:peru}
\end{figure}

We further test our method on the multi-fidelity climate dataset provided by Hosking \citep{Hosking2020}. The dataset includes low-fidelity and high-fidelity climate model temperature simulations over a region in Peru. The left part of Figure \ref{fig:peru} shows the region of interest.

\textbf{Dataset.} The low-fidelity data is generated by low-fidelity Global Climate Model with spatial resolution $14 \times 14$ \citep{noaa2009}. The high-fidelity data is generated by high-fidelity Regional Climate Model \citep{armstrong2019reassessing} with spatial resolution $87 \times 87$. The example is shown in Figure \ref{fig:peru}. Both include monthly data from 1980 to 2018 over the same region (latitude range: $(-7.5,-10.7)$, longitude range: $(280.5,283.7)$).

The task is to use $6$ month data as input to generate the next $6$ month predictions as output. We randomly split $119$ scenarios for training candidate set, $50$ scenarios for validation set, and $50$ scenarios for the test set at both fidelity level. In the nested data set case, we first randomly select $87$ scenarios from the training candidate set as the training set at low-fidelity level, then randomly select $32$ scenarios from them as the training set at high-fidelity level. In the non-nested data set case, we randomly split $87$ scenarios as the training set at low-fidelity level and $32$ scenarios as the training set at high-fidelity level. The validation and test set are both at high-fidelity level.

\begin{figure*}[t]
    \centering
    \includegraphics[width=0.95\linewidth]{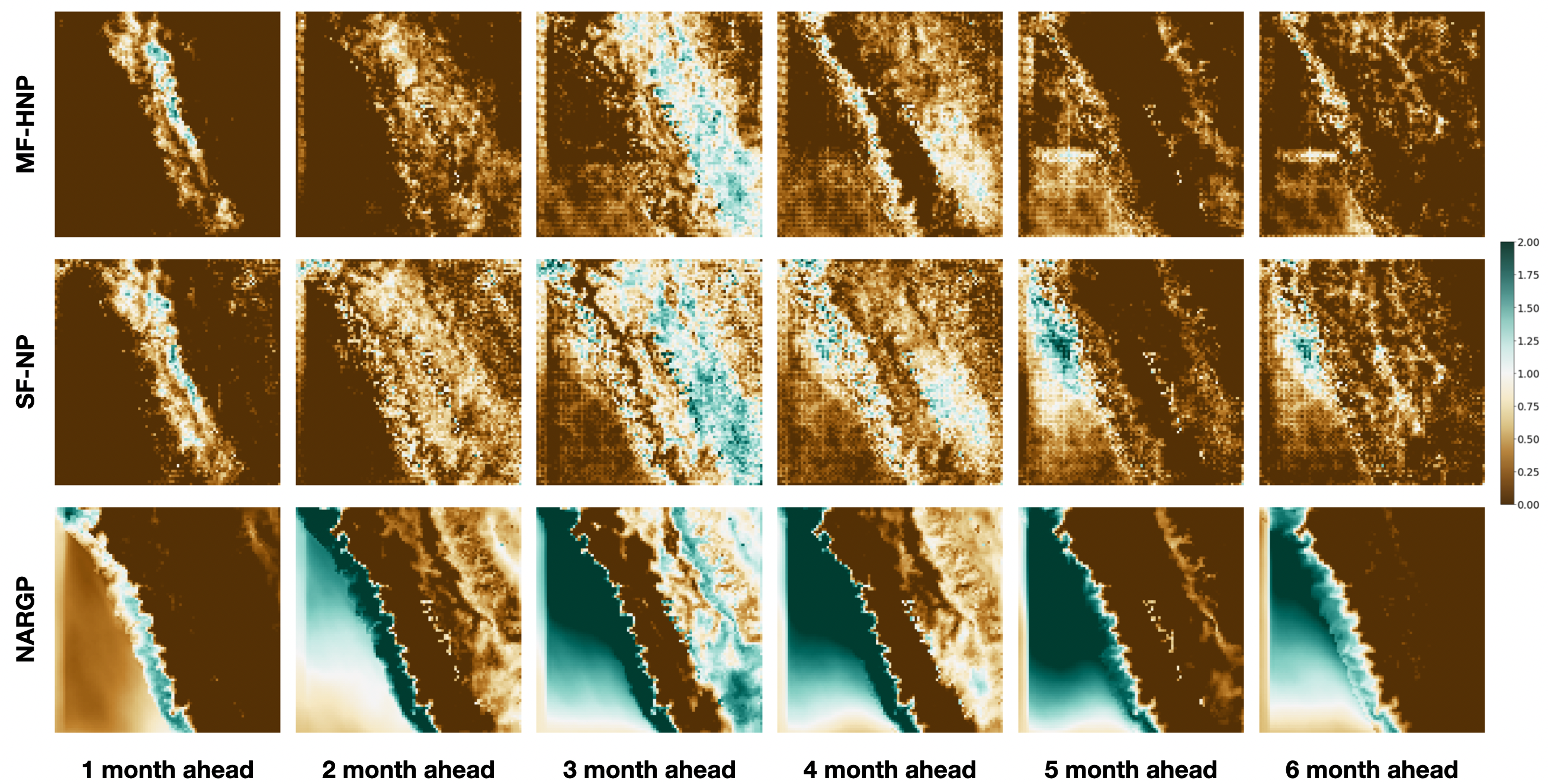}
    \caption{\ours{} vs. SF-NP vs. NARGP for 6 month ahead temperature prediction residual.}
    \label{fig:fig5}
\end{figure*}

\begin{table*}[h]
\caption{Prediction performance comparison on climate data sets.}
\vskip -2mm
\label{tb:climate-table}
% \vskip 0.15in
\begin{center}
\begin{sc}
\begin{tabular}{c|c|c|c|c}
\toprule
Method & MAE (nested) $\downarrow$ & NLL (nested) $\downarrow$ & MAE (non-nested) $\downarrow$ & NLL (non-nested) $\downarrow$\\
\midrule
SF-GP & $0.91\pm0.365$ & $2.288\pm0.004$ &  $0.91\pm0.365$ & $2.288\pm0.004$\\ 
NARGP & $0.91\pm0.365$ & $2.3\pm0.006$ & $\times$  & $\times$ \\ 
SF-NP-MA  & $0.778\pm0.01$ & $\mathbf{1.489}\pm0.026$ & $0.778\pm0.01$ & $\mathbf{1.489}\pm0.026$ \\ 
MF-NP-MA  & $0.902\pm0.005$ & $1.889\pm0.012$ & $\times$  & $\times$ \\
MF-HNP(mean)-MA  & $0.765\pm0.004$ & $1.535\pm0.059$  & $0.788\pm0.029$ & $1.666\pm0.174$\\
MF-HNP(mean,std)-MA  & $0.773\pm0.011$ & $1.592\pm0.057$ & $0.768\pm0.027$ & $1.607\pm0.089$\\
MF-HNP(as)-MA  & $0.758\pm0.024$ & $1.578\pm0.079$  & $0.769\pm0.02$ & $1.594\pm0.098$\\
SF-NP-BA  & $0.751\pm0.052$ & $1.546\pm0.133$  & $0.751\pm0.052$ & $1.546\pm0.133$ \\ 
MF-NP-BA  & $0.954\pm0.019$ & $1.909\pm0.028$  & $\times$  & $\times$ \\
MF-HNP(mean)-BA  & $0.706\pm0.049$ & $1.549\pm0.164$  & $0.714\pm0.027$ & $1.58\pm0.061$\\
MF-HNP(mean,std)-BA  & $0.717\pm0.045$ & $1.606\pm0.106$  & $0.695\pm0.03$ & $1.548\pm0.068$\\
MF-HNP(as)-BA  & $\mathbf{0.678}\pm0.026$ & $1.506\pm0.027$  & $\mathbf{0.68}\pm0.009$ & $1.58\pm0.012$\\
\bottomrule
\end{tabular}
\end{sc}
\end{center}
\vskip -0.1in
\end{table*}

\textbf{Performance Analysis.}
Table \ref{tb:climate-table} compares the prediction performance for 2 GP methods and 10 NP methods to predict the next 6 months temperature based on the past 6 months temperature data. The performance is reported in MAE and NLL. The results of this task are consistent with what we found in AS-SIR infection prediction task. \ours{} has significantly better performance compared with either GP or NP baselines. But this time \ours{}(MC)-BA is the most accurate one with or without a nested data structure. Considering both MAE and NLL, we still recommend using \ours{}(MC)-BA and \ours{}(MEAN)-BA. 

Figure \ref{fig:fig5} is the visualization of predictions among the best \ours{} variant, GP and NP baselines on a randomly selected scenario in the test set. To highlight the performance difference, we visualize the residual between the predictions and the truth from 1 to 6 months ahead predictions. Higher value means lower accuracy. It can be found that \ours{} outperforms all the baselines for the predictions for each month.

\section{Conclusion \& Limitation}
We propose  Multi-Fidelity Hierarchical Neural Process (\ours{}), the first \textit{unified} framework for scalable multi-fidelity surrogate modeling in the neural processes family. Our model is more flexible and scalable compared with existing multi-fidelity modeling approaches. Specifically, it no longer requires a nested data structure for training and supports varying input and output dimensions at different fidelity levels. Moreover, the latent variables introduce  conditional independence for  different fidelity levels, which  alleviates the error propagation issue and improves the accuracy and uncertainty estimation performance. We demonstrate the superiority of our method on two real-world large-scale multi-fidelity applications: age-stratified epidemiology modeling and temperature outputs from different climate models. 

Regarding future work, it is natural to extend our  multi-fidelity Hierarchical Neural Process to active learning setup. Instead of passively training the neural processes, we can proactively query the simulator, gather training data, and incrementally improve the surrogate model performance.

\section*{Acknowledgement}
This work was supported in part by U.S. Department Of Energy, Office of Science under Award \#DE-SC0022255, U. S. Army Research Office under Grant W911NF-20-1-0334, Facebook Data Science Research Awards, AWS Machine Learning Research Award, Google Faculty Award, NSF Grants \#2037745, NSF-SCALE MoDL-2134209, and NSF-CCF-2112665 (TILOS). D.W. acknowledges support from the HDSI Ph.D. Fellowship. M.C. and A.V. acknowledge support from grants HHS/CDC 5U01IP0001137 and HHS/CDC 6U01IP001137. The findings and conclusions in this study are those of the authors and do not necessarily represent the official position of the funding agencies, the National Institutes of Health, or the U.S. Department of Health and Human Services.
\bibliographystyle{ACM-Reference-Format}
\bibliography{ref}

\clearpage
% \onecolumn
\appendix
\section{Neural Processs Baselines}
\label{Appendix:np_baseline}
\subsection{Single-Fidelity Neural Processes (SF-NP).}
A simple way to apply NP to the multi-fidelity problem is to train NP only using the data at high-fidelity level only assuming it is not correlated with the data at the low-fidelity level. We name it as Single-Fidelity Neural Processes baseline (SF-NP). During the training process, the high-level training data can be randomly split into context set $\mathcal{D}^c_h$ and target set $\mathcal{D}^t_h$. We use the corresponding evidence lower bound (ELBO) as the training loss function:

% \begin{align}
% \begin{split}
%     & \log p(y^h_t|x^h_t,x^h_c,y^h_c,\theta) \geq\\ 
%     & \mathbb{E}_{q_\phi(z|x^h_c,y^h_c,x^h_t,y^h_t)} \big[  \log p(y^h_t|z, x^h_t,\theta)\big] - \\
%     & \text{KL}\big( q_\phi(z| x^h_c,y^h_c)\| q_\phi(z|x^h_c,y^h_c,x^h_t,y^h_t ) \big)
%     \label{eqn:sfnp}
% \end{split}
% \end{align}

\begin{align}
% \begin{split}
    & \log p(y^t_{h,1:M}|x^t_{h,1:M},\mathcal{D}^c_h,\theta) \geq  \nonumber\\ 
    & \mathbb{E}_{q_\phi(z|\mathcal{D}^c_h \cup \mathcal{D}^t_h)} \big[ \sum_{m=1}^M\log p(y^t_{h,m}|z, x^t_{h,m},\theta)+log\frac{q_\phi(z|\mathcal{D}^c_h)}{q_\phi(z|\mathcal{D}^c_h \cup \mathcal{D}^t_h)}\big] \nonumber
    % \label{eqn:elbo_sfnp}
% \end{split}
\end{align}

where $p(\theta)$ is a decoder in a neural network and $q_\phi$ indicates a encoder to infer the latent variable $z$.

% \allen{how to generate predictions, conditioned on training set}

\subsection{Multi-Fidelity Neural Processes (MF-NP).} 
Multi-Fidelity Neural Processes (MF-NP) \cite{wang2020mfpc} assume a comprehensive correlation between multi-fidelity models $y_h$ and $y_l$ can be represented as:
\begin{align}
% \begin{split}
    y_h(x) = \mathcal{G}(y_l(x)) + \delta(x),  \nonumber
    % \label{eqn:mfnp_gen}
% \end{split}
\end{align}
where $\mathcal{G}$ is a nonlinear function mapping the low-fidelity data to high-fidelity data, and $\delta(x)$ is space dependent bias between fidelity levels.
To train MF-NP model, we take data pairs $(x,y_l(x))$ as the input to predict the corresponding $y_h(x)$. The corresponding context sets $\mathcal{D}_l^c \equiv \{x^c_{h,n},{y^c_{l,n}}, {y^c_{h,n}}\}_{n=1}^{N_l}$ and target sets $\mathcal{D}_l^t \equiv \{x^t_{h,m},{y^t_{l,m}},{y^t_{h,n}}\}_{m=1}^{M_l}$. The ELBO for the training process is:

% \begin{align}
% \begin{split}
%     &\log p(y^h_t|x^h_t,y^l_t,x^h_c,y^l_c,y^h_c,\theta) \geq\\ &\mathbb{E}_{q(z|x^h_c,y^l_c,y^h_c,x^h_t,y^l_t,y^h_t)} \big[
%     \log p(y^h_t|z, x^h_t, y^l_t, \theta)\big] - \\
%     & \text{KL}\big( q_\phi(z| x^h_c,y^l_c,y^h_c)\| q_\phi(z|x^h_c,y^l_c,y^h_c,x^h_t,y^l_t,y^h_t) \big)
%     \label{eqn:mfnp}
% \end{split}
% \end{align}

\begin{align}
% \begin{split}
    & \log p(y^t_{h,1:M}|x^t_{h,1:M},y^t_{l,1:M},\mathcal{D}^c_h,\theta) \geq \nonumber\\ 
    & \mathbb{E}_{q_\phi(z|\mathcal{D}^c_h \cup \mathcal{D}^t_h)} \big[ \sum_{m=1}^M\log p(y^t_{h,m}|z, x^t_{h,m},y^t_{l,m},\theta)+ \nonumber\\ 
    & log\frac{q_\phi(z|\mathcal{D}^c_h)}{q_\phi(z|\mathcal{D}^c_h \cup \mathcal{D}^t_h)}\big]  \nonumber
    % \label{eqn:elbo_mfnp}
% \end{split}
\end{align}

Since this method requires $(x,y_l(x),y_h(x))$ for input and output, it can not fully utilize the training data at low-fidelity level which $y_h(x)$ is unknown.  Furthermore, MF-NP requires a nested data structure, which means the training inputs of high-fidelity level need to be a subset of the training inputs of low-fidelity level. On the contrary, if  the training inputs at the different fidelity level are disjoint, no data set can be used for training.

\section{Experiment Details}
For GP baselines, we use RBF kernels. The optimal learning rate is $5e^{-2}$ for both AS-SIR and climate modeling tasks. We train $2000$ epochs with patience equal to $100$ to ensure convergence. 
For NP baselines and our proposed \ours{} model, the hyperparameters can be found in Table \ref{tb:hyper-table}.

\begin{table}
\caption{Hyperparameters for NP baselines and our proposed \ours{} model, including learning rate, batch size, and patience.}
\label{tb:hyper-table}
\begin{center}
\begin{sc}
\begin{tabular}{c|c|c|c}
\toprule
 & learning rate & batch size & patience \\ 
\midrule
AS-SIR & $1e^{-3}$          & 128                 & 1000     \\ 
climate         & $5e^{-3}$          & 32             & 250      \\
\bottomrule
\end{tabular}
\end{sc}
\end{center}
\end{table}
\end{document}